\documentclass[letterpaper,10 pt,journal,twoside]{IEEEtran}
\usepackage{amsmath}
\usepackage{algorithmic}
\usepackage{array}
\usepackage{stfloats}
\usepackage{url}
\usepackage{cite}
\usepackage{graphicx}
\usepackage{mathtools}
\usepackage{threeparttable}
\usepackage{amssymb}
\usepackage{bm}
\usepackage{float}
\usepackage{subfigure}
\usepackage{gensymb}
\usepackage{textcomp}
\usepackage{wasysym}
\usepackage{xcolor,soul}
\usepackage{hyperref}
\usepackage{booktabs}
\usepackage{multirow}
\usepackage{colortbl,hhline}
\usepackage{mathtools}
\usepackage[ruled,linesnumbered,vlined]{algorithm2e} 

\usepackage{tikz,xcolor,hyperref}

\definecolor{lime}{HTML}{A6CE39}
\DeclareRobustCommand{\orcidicon}{%
	\begin{tikzpicture}
		\draw[lime, fill=lime] (0,0) 
		circle [radius=0.16] 
		node[white] {{\fontfamily{qag}\selectfont \tiny ID}};
		\draw[white, fill=white] (-0.0625,0.095) 
		circle [radius=0.007];
	\end{tikzpicture}
	\hspace{-2mm}
}

\foreach \x in {A, ..., Z}{%
	\expandafter\xdef\csname orcid\x\endcsname{\noexpand\href{https://orcid.org/\csname orcidauthor\x\endcsname}{\noexpand\orcidicon}}
}

\hypersetup{
	colorlinks   = true,
	urlcolor     = blue,
	linkcolor    = blue,
	citecolor    = magenta
}

\renewcommand{\baselinestretch}{0.98}

\begin{document}

\title{\huge Constrained Motion Planning of A Cable-Driven Soft Robot With Compressible Curvature Modeling}

\author{Jiewen Lai,~Bo Lu, Qingxiang Zhao, and Henry K. Chu
	
\thanks{This work was supported in part by the Departmental Grant 4-RKCM of ME-HKPolyU. \textit{(Corresponding author: Henry K. Chu.)}}

\thanks{J. Lai, Q. Zhao, and H. K. Chu are with the Department of Mechanical Engineering, The Hong Kong Polytechnic University, Kowloon, Hong Kong SAR, China. (email: \href{mailto:jw.lai@connect.polyu.hk}{jw.lai@connect.polyu.hk}; \href{mailto:qingxiang.zhao@connect.polyu.hk}{qingxiang.zhao@connect.polyu.hk}; \href{mailto:henry.chu@polyu.edu.hk}{henry.chu@polyu.edu.hk}) } 

\thanks{B. Lu is with the School of Mechanical and Electrical Engineering, Robotics and Micro-systems Center, Soochow University, China. (email: \href{mailto: blu@suda.edu.cn}{blu@suda.edu.cn})}

\thanks{Video demo: \href{https://youtu.be/Zb3cA3hTvKg}{https://youtu.be/Zb3cA3hTvKg}}


}



\maketitle

\begin{abstract}
A cable-driven soft-bodied robot with redundancy can perform the tip trajectory tracking task and in the meanwhile fulfill some extra constraints, such as tracking with a designated tip orientation, or avoiding obstacles in the environment. These constraints require proper motion planning of soft material-based body that can be axially compressed. In this letter, we derived the compressible curvature kinematics of a cable-driven soft robot which takes the undesirable axial compression that caused by the cable-driven mechanism into account. The motion planning of the soft robot for tip trajectory tracking tasks in constrained conditions, including fixed orientation end-effector and manipulator--obstacle collision avoidance, have been investigated. The inverse solution of cable actuation was formulated as a damped least-square optimization problem and iteratively computed off-line. The performance of path and trajectory tracking and the obedience to constraints were evaluated via the simulation we made open-source, as well as the prototype experiments. The method can be generalized to the similar multisegment cable-driven soft robotic systems by customizing the robot parameters for the prior motion planning in tip trajectory following tasks.
\end{abstract}

\IEEEpeerreviewmaketitle

\section{Introduction}
\label{introduction}
\IEEEPARstart{S}{oft} robots are primarily composed of materials with low Young's moduli that are comparable with the biological materials like muscles \cite{rus2015design,cianchetti2014soft}. They can be potentially used in many applications, such as robot-assisted minimally invasive surgery \cite{cianchetti2014soft,runciman2019soft,burgner2015continuum,unpublish} and laser steering \cite{mo2021control}. Using a robot tip to track a designated trajectory poses an essential scene of automation. The field of conducting trajectory tracking using rigid-bodied robots has been well-explored, while challenges remain for the soft-bodied robots, especially when they are redundant. The elasticity of the soft materials, as well as the way of actuation, would bring complex variation to the robot modeling.

For a \textit{Cable-Driven Soft (Continuum) Robot} (CDSR) which requires at least 3 actuation cables to achieve spatial manipulability, one can adopt the geometrically-derived piecewise constant curvature (PCC) approach to obtain the closed-form solutions for modeling \cite{webster2010,camarillo2008mechanics,xu2008investigation,della2020improved}. 
If assuming the arc length (robot's ``backbone'') to be constant, the PCC would simplify a robot segment to a 2-DOF module. It has been experimentally validated that the neutral axis of a continuum segment would conditionally perform as a constant-curvature arc \cite{xu2008investigation}.

{However, the simplification or approximation of constant arc length in solving the modeling equations may not precisely reveal the robot configuration, especially when the axial compression of the soft body is significant enough to vary the body stiffness along with different directions} \cite{9427246}. Some groups provided a general approach to update the arc length with respect to the bending \cite{della2020improved}. Other model-based methods with better applicability have also been proposed \cite{camarillo2008mechanics,camarillo2009configuration,rone2014mechanics,gonthina2020mechanics,boyer2020dynamics}. {Particularly, the inevitable axial compression of the cable-driven continuum robots has been investigated and modeled in \cite{camarillo2008mechanics} and \cite{camarillo2009configuration} using a mechanic-based method. Based on a series of proof-of-concept verification in 2-D, the actuation-associated axial strain has been studied with the cable decoupling being considered at the system level. This could be further introduced to the 3-D scenarios with the constrained motion being assigned.}

The other thing is that many soft robots reported in the literature exclude the control of tip orientation, i.e., only the tip position is considered \cite{qi2016kinematic,xu2020adaptive}. One of the major reasons is that the adoption of the PCC model simplifies the controllable DOFs, requiring a soft manipulator to have at least three segments to have full control in the task space \cite{godage2016dynamics}. But prototyping a CDSR with three or even more segments could be challenging due to the densely distributed cables, especially when it comes to the millimeter-grade, not to mention the complicated coupling among the segments. 


Moreover, a redundant robot is also capable of providing alternative motions for the designated tip position, and these motions can lead to collision-free control in the obstructed environment. Related work have been done in terms of simulation \cite{xiao2010real,ataka2016real} and 2-D prototype experiments \cite{roesthuis2016steering,marchese2014whole,marchese2016dynamics}. Collision-free motion planning for a redundant CDSR under the constrained conditions remains an important topic in the field. On one hand, the cable-driven mechanism benefits the miniaturization of a spatially maneuverable soft robot. On the other hand, proper motion planning would gift the redundant soft robots the ability to work in constrained conditions, such as the task-based manipulation and manipulator-obstacle collision avoidance, both of which might interest the physicians.

In this letter, we {explicitly} derive the kinematic model for a multi-segment cable-driven soft robot on a linear stage (i.e., $3n+1$ DOFs, where $n$ denotes the number of soft segments with three cables for actuation) with improved accuracy by accounting for the passive axial compression which is undesirably caused by the actuation and the coupling effect between segments. The conventional PCC provides the framework for the model-based control of continuum robots, but the applicability is not good enough for CDSRs where the multiple cable actuation for desired bending would at the same time introduce undesirable axial compression and coupling. Our improved model offers a higher accuracy and will be more realistic in the task space with the above-mentioned condition considered. In addition, a multi-segment CDSR simulator was developed to illustrate the robot performance under different constrained conditions, including (i) fixing the orientation of the tip, and (ii) avoiding obstacles in tip path following. The improved model and motion planning method were also examined through an experimental platform with 2 segments to evaluate the actual performance and compare with the simulation results.

\section{Compressible Curvature Model}
\label{ncc}
\subsection{Mechanics of A Single Soft Segment}
A continuum manipulator is usually designed with a flexible backbone along the neutral line to support the equidistantly distributed discs. The flexible backbone has a high stiffness $K_{a}$ along its axial spine, and relatively low stiffness $K_{b}$ in its lateral direction, so that it provide an ideal ``bending'' deformation. However, for a soft material-based robot without a backbone, the axial stiffness $K_{a}$ can be small, and neglecting its effect could cause a perceptible error in the task space, given that most of the maneuverable soft robots are extrinsically driven. Therefore, we take the relationship between cable-driven mechanism and its undesirable axial compression into the modeling consideration.
\begin{figure}[t]
	\centering
	\includegraphics[width = 3.2in]{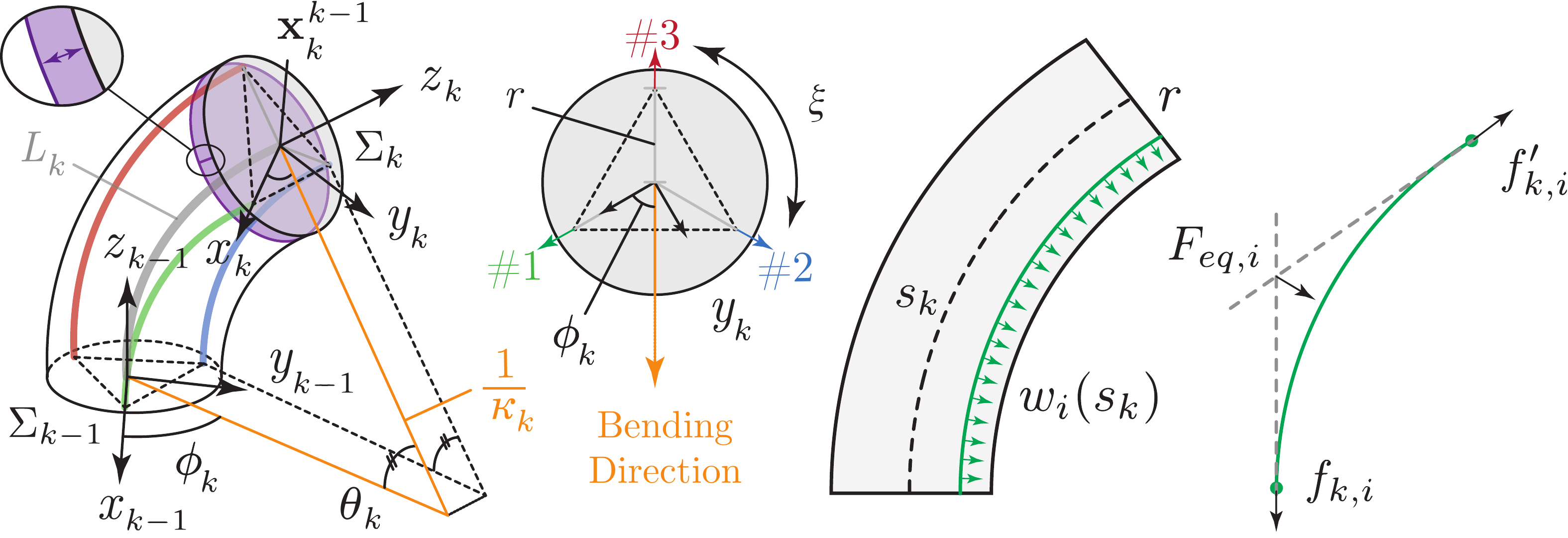}
	\caption{Sketch of a CDSR. The axial compression is colored in purple. The sketch at the right is partially modified from \cite{camarillo2008mechanics}).}
	\label{segment}
\end{figure}

Leveraging from the PCC model in \cite{webster2010}, for a multi{\tiny }segment soft robot with $n$ segments, the $k$-th soft segment ($k\leq n$) can be geometrically parameterized in the configuration space by $\bm{\psi}_k=[\theta_k, \phi_k, \kappa_k]^{{\top}}$ where $\theta_k\in\left[0,\theta_{max}\right] $ is the angle of bending, $\phi_k\in\left[ -\pi,\pi\right) $ is the angle of bending direction, and $\kappa_k$ is the bending curvature related to the initial undeformed length of the segment $L_k$ (therefore, time-invariant) and $\kappa_k=\theta_k/ L_k$. Thus, the position of the $k$-th segment with respect to its base frame $\Sigma_{k-1}$ can be expressed by
\begin{equation}
	\normalsize
	\label{position}
	\mathbf{x}_{k}^{k-1} = \begin{bmatrix}
		{x}_k\\{y}_k\\{z}_k
	\end{bmatrix}=\frac{1}{\kappa_k}
	\begin{bmatrix}
		\cos{\phi_k}\left( 1-\cos{\theta_k}\right)\\
		\sin{\phi_k}\left( 1-\cos{\theta_k}\right)\\
		\sin{\theta_k}
	\end{bmatrix}.
\end{equation}
As depicted in Fig. \ref{segment}, the configuration of the $k$-th segment $\bm{\psi}_k$ can be derived from the cable actuation as \cite{webster2010}
\begin{equation}
	\normalsize
	\label{theta}
	\theta_k(\bm{q})\!=\!\frac{2\sqrt{q_{k,1}^2\!+\!q_{k,2}^2\!+\!q_{k,3}^2\!-\!q_{k,1}q_{k,2}\!-\!q_{k,1}q_{k,3}\!-\!q_{k,2}q_{k,3}}}{3r},
\end{equation}
\begin{equation}
	\normalsize
	\label{phi}
	\phi_k(\bm{q}) \!=\!\mathrm{atan2}\left(3\left(q_{k,2}\!-\!q_{k,3} \right) ,\sqrt{3}\left( q_{k,2}\!+\!q_{k,3}\!-\!2q_{k,1}\right) \right),
\end{equation}
\begin{equation}
	\normalsize
	\label{kappa}
	\kappa_k(\bm{q}) \!=\! \frac{2\sqrt{q_{k,1}^2\!+\!q_{k,2}^2\!+\!q_{k,3}^2\!-\!q_{k,1}q_{k,2}\!-\!q_{k,1}q_{k,3}\!-\!q_{k,2}q_{k,3}}}{3r/{L_k}},
\end{equation}
where $r$ denotes the (constant) symmetric distance between the center of the cross-section and the cable channel, and $q_{k,i} \subseteq \bm{q}_k$ denotes the actuation-by-length of the $i$-th cable of $k$-th segment, given by the subtraction of the in-body cable length $L_{k,i}$ and the length of the neutral line $L_k$ as
\begin{equation}
	\normalsize
	\label{q}
	q_{k,i}=L_{k,i}-L_k={-\theta_k r\cos{\left( \phi_k+\left(i-1 \right)\xi\right)}},
\end{equation}
where $i\in\left\lbrace1,2,3 \right\rbrace$ and $\xi=\frac{2\pi}{3}$. Thus, the joint variables of the $k$-th segment expressed in the linear displacements of the actuation cables can be written in the form of $\bm{q}_k=[q_{k,1},q_{k,2},q_{k,3}]^{\top}\in\mathcal{Q}^3$, with $\mathcal{Q}^3$ being an admissible space on which the unconstrained joint variables can evolve. Note that $\mathcal{Q}^3\not\subset\mathbb{R}^3$ since the parameter $\theta_k\in[0,\theta_{max}]$ is bounded. The above equations (\ref{position})--(\ref{q}) conclude the PCC model of a continuum segment, of which the neutral line $L_k$ is assumed to be constant. However, for a compressible soft segment, the segment length $s_k$ is a variable related to the tensile force of cables---within the elastic limit, a shorter segment would be expected under a higher tensile force. Note that $s_k\leq L_k$. It results in a compressible curvature arc. The cable length variables can be mapped with the tensile force variables under several assumptions, including (i) the neutral line deforms as a circular arc, (ii) the soft segment demonstrates a linear elasticity under tensile force, (iii) the friction between the cable and channel is omitted, (iv) the gravitational potential energy of the elastomer is neglected \cite{xu2008investigation} (see supplementary).

For a multisegment soft robot with $n$ segments coupled together, as a result of the accumulative tensile force, the axial strain of the ($k-1$)-th segment would be larger than that of the $k$-th segment. In the local $k$-th segment, we assume the tensile force of the cables as $\bm{f}_k=[f_{k,1},f_{k,2},f_{k,3}]^{\top}$. When cables are actuated and at the equilibrium, a suitable candidate function $\bm{f}_k:\mathcal{Q}^3\mapsto\mathbb{R}^3$ that maps to the joint variables to a force vector that satisfies the quasi-static equilibrium for all $q_{k,i}\in\mathcal{Q}^3$ can be concluded, such that
\begin{equation}
	\normalsize
	\label{phikf}
	\phi_k\left(\bm{f} \right)  =\mathrm{atan2}\left(3\left(f_{k,2}-f_{k,3} \right) ,\sqrt{3}\left( f_{k,2}+f_{k,3}-2f_{k,1}\right) \right).
\end{equation}
In a quasi-static state, as shown in Fig. {\ref{segment}}, we can write the static equilibrium equation for a single cable as \cite{camarillo2008mechanics}
\begin{equation}
	\mathbf{F}_{k,i}=\bm{F}_{eq,i}+\bm{f}_{k,i}+\bm{f}^{\prime}_{k,i}=\mathbf{0},
\end{equation}
where $\bm{f}_{k,i}$ and $\bm{f}^{\prime}_{k,i}$ are the cable tensile and the reaction force perpendicular to the tip plane, respectively. Note that, different from the unbolded $f_{k,i}$, the bolded $\bm{f}_{k,i}$ represents the vector of tensile. $\bm{F}_{eq,i}\in\mathbb{R}^3$ denotes the resultant equilibrium force that bends the segment, which can be derived by taking the integration over the time-variant arc length $s_k$ as
\begin{equation}
	\normalsize
	\label{feqi}
	\begin{aligned}
	\bm{F}_{eq,i}&=\int_{0}^{s_k}\bm{w}_i\left(\sigma_k \right) \mathrm{d}\sigma_k=\kappa_k^{-1}\int_{0}^{\theta_k}\bm{w}_i\left(\sigma_k \right)\mathrm{d}\sigma^{\left[ \theta\right] }_k \\
	&=\kappa_k^{-1}\int_{0}^{\theta_k}\!\mathbf{R}_k^{k-1}\cdot
	\begin{bmatrix}
		-f_{k,i}\kappa_k\cos\left(\phi_k\!+\!\left( i\!-\!1\right)\xi\right)\\
		-f_{k,i}\kappa_k\sin\left(\phi_k\!+\!\left( i\!-\!1\right)\xi\right)\\
		0
	\end{bmatrix}\!{\mathrm{d}\sigma^{\left[ \theta\right] }_k},
	\end{aligned}
\end{equation}
where $\bm{w}_i\left(\sigma_k \right)$ denotes the distributed load in an infinitesimal cable length of $\sigma_k$, with $\mathrm{d}\sigma_k=\kappa_k^{-1}\mathrm{d}\sigma^{\left[ \theta\right] }_k$, and $\mathbf{R}_k^{k-1}$ is the rotation matrix that defines the $k$-th frame orientation w.r.t. that of the $(k-1)$-th frame as
\begin{equation}
	\normalsize
	{\mathbf{R}}_k^{k-1} = \mathrm{Rot}\left( \mathbf{\bar z}_{k-1},{\phi}_k\right) \cdot \mathrm{Rot}\left( \mathbf{\bar y}_{k-1},{\theta}_k\right) \cdot\mathrm{Rot}\left( \mathbf{\bar z}_{k},-{\phi}_k\right),
\end{equation}
where the operator $\mathrm{Rot}(\mathbf{\bar u}_{k},\delta)\in SO\left(3 \right) $ is the rotation matrix of rotating angle $\delta$ along the unit axis $\mathbf{\bar u}_{k}$ of the $k$-th segment. 

As the cables are equally distributed, based on Eq. \eqref{feqi}, we take the triple product of them and find out that
\begin{equation}
	\bm{F}_{eq,1}^{\top}\cdot\left(\bm{F}_{eq,2}\times \bm{F}_{eq,3} \right)=\mathbf{0}, 
\end{equation}
meaning that the three resultant equilibrium forces are \textit{coplanar} vectors (abbreviation for $\mathrm{s}=\sin$ and $\mathrm{c}=\cos$). Given the fact that $\bm{F}_{eq,i}$ is a function of $\bm{f}_{k,i}$, we can conveniently decouple the tensiles. As the soft segment bends, the resultant of torque acting on the tip of the $k$-th segment, ${\tau}_k$, can be derived from the sum of cable tensile as
\begin{equation}
	\label{tau}
	\normalsize
	\begin{aligned}
		{\tau}_k=\sum_{i=1}^{3}\left( \bm{r}_i\times\bm{f}_{k,i}\right) 
		={r} \sum_{i=1}^{3} \left( f_{k,i}\cos\left(\phi_k+\left( i-1\right)\xi\right) \right),
	\end{aligned}
\end{equation}
where $\times$ denotes the cross-product operation because there are $i$ different cables, and $\|\bm{r}_i\|=r$, with $\bm{r}_i$ being the concerning vector from the neutral line to the $i$-th cable in a distance of $r$. From (\ref{phikf}), one can obtain $\cos\phi_k(\bm{f})$ and $\sin\phi_k(\bm{f})$ (see supplementary), 
such that (\ref{tau}) can be rewritten as
\begin{equation}
	\normalsize
	\tau_k = -r\mathcal{F}_k,
\end{equation}
where
	$\mathcal{F}_k\!=\!\sqrt{f_{k,1}^2\!+\!f_{k,2}^2\!+\!f_{k,3}^2\!-\!f_{k,1}f_{k,2}\!-\!f_{k,1}f_{k,3}\!-\!f_{k,2}f_{k,3}}$.

{In this model, we consider the cross-sectional area as $\mathcal{A}_k=\pi\left(r_{o,k}^2-r_{i,k}^2 \right)$ where both radii are variables subjected to the compression of the soft body, and the segment length can be derived as
\begin{equation}
	s_{k}\left(\bm{f} \right) =L_k\left(1+\frac{\sum_{j=k}^{n}\sum_{i=1}^{3}f_{j,i}}{E\mathcal{A}_k } \right) 
\end{equation}
which in fact defines a variable axial stiffness. The variable radii are given by
\begin{equation}
	\label{r_var}
	\begin{cases}
		 r_{o,k}\left(\bm{f} \right)  =\displaystyle r_{o}\left(1-\frac{\nu{\sum_{j=k}^{n}\sum_{i=1}^{3}f_{j,i}}}{E\pi\left( r_{o,k}^2-r_{i,k}^2\right) } \right) \\
		 r_{i,k}\left(\bm{f} \right) =\displaystyle r_{i}\left(1-\frac{\nu{\sum_{j=k}^{n}\sum_{i=1}^{3}f_{j,i}}}{E\pi\left( r_{o,k}^2-r_{i,k}^2\right) } \right),
	\end{cases}
\end{equation}
where $r_o$ and $r_i$ are the original radii of the soft segment, and $\nu=0.45$ is the Poisson's ratio of the soft body. The derivation of Eq. \eqref{r_var} is available in the supplemented page and codes. {When the robot is compact with a relatively large aspect ratio ($\frac{r_o}{L_k}$), like the robot adopted in our paper, the effect on the radial change may not be obvious. Nevertheless, robots made from soft materials with a high Poisson’s ratio, low elastic modulus, and a long segment length would see the difference.}}

The local bending angle can be derived as the free-end slope of a cantilever beam under a couple moment of $M_k$ by ${\mathrm{d}\theta_k}/{\mathrm{d}s}=M_{k}/{E I}$, which yields
\begin{equation}
	\normalsize
	\label{thetakf}
	\theta_k\left(\bm{f} \right) =\int_{0}^{s_k}{\frac{M_{k}}{EI}}\mathrm{d}s=\frac{r\mathcal{F}_k s_k}{K_b}=\frac{\mathcal{M}_k}{K_{T,k}},
\end{equation}
where $K_b = EI$ is the flexural rigidity of the structure, $M_{k}=|\tau_k|$ is the {local} bending moment of the $k$-th segment, {$\mathcal{M}_k$ is the bending moment of the $k$-th segment subjected to the variable length, and particularly,}
\begin{equation}
	\normalsize
	\label{kb/s}
	K_{T,k}\left(s_k\right)  =\frac{K_b}{s_k }=\frac{EI}{s_k}=\frac{E\pi\left(r_{o,k}^4-r_{i,k}^4 \right) }{4s_k }
\end{equation}
denotes the bending stiffness of the $k$-th segment, {which can vary dramatically for a soft-bodied robot.} And thus, the curvature related to the cable tensile can be computed as
\begin{equation}
	\normalsize
	\label{kappakf}
\kappa_k\left(\bm{f} \right)=\frac{\theta_k\left(\bm{f} \right) }{s_k\left( \bm{f}\right)}.
\end{equation}
{Eq. \eqref{kb/s} indicates the fact that the segment compression would result in a variable bending stiffness locally, and the proximity toward the base implies an even stiffer case. Therefore, for a multisegment soft robot with coupled cables, the local bending stiffness of each segment shall be different and relevant if it is not being the distal-most. Instead of terming the local bending stiffness independently \cite{camarillo2009configuration,rone2014mechanics,gonthina2020mechanics}, we formulate the variable stiffnesses that are responsive to the axial compression, therefore, the cable actuation. The formulation suggests that the undesirable axial compression will not only affect the tip positioning, but also the necessary inverse solution in reality.} {A numerical verification will be provided in Sec. \ref{Model Verification}.}

\subsection{Mechanics of A Coupled Soft Segment}
Since the cables of the distal segment would pass through the channels of the proximal segments, the coupling effects shall be considered. In this subsection, we use a hat sign to represent the coupled configurations of the relative proximal segments, namely, $\widehat{\bm{\psi}}_k=[\widehat{\theta}_k,\widehat{\phi}_k,\widehat{\kappa}_k]^{\top}$, where $k\neq n$.

The bending angle of the far-most $n$-th segment, $\widehat{\theta}_n$, is only related to the local tensile $\bm{f}_{n}$. Therefore, its configuration can be obtained using (\ref{phikf}), (\ref{thetakf}), and (\ref{kappakf}). However, the bending and twisting of the ($n-1$)-th segment is subjected to the local actuation $\bm{q}_{n-1}$ as well as the successive actuation $\bm{q}_{n}$. The manipulation of the successive segment would generate an additional moment that couples the configuration of its previous segment backward \cite{camarillo2009configuration}, denoted by
\begin{equation}
	\normalsize
	\label{m+m}
	\widehat{\mathcal{M}}_{n-1} = \mathcal{M}_{n-1} + \widehat{\mathcal{M}}_{n},
\end{equation}
where $\widehat{\mathcal{M}}_{n-1}$ denotes the incorporated moment acting on the ($n-1$)-th segment, and $\widehat{\mathcal{M}}_{n}\equiv{\mathcal{M}}_{n}$. Eq. \eqref{m+m} can be trigonometrically solved by giving arbitrary local directions \cite{gonthina2020mechanics}. 
After reformulation, the general coupled configuration can be solved as
\begin{equation}
	\normalsize
	\begin{aligned}
	\widehat{\theta}_{n-1}=&\frac{1}{K_{T,n-1}}\left({\mathcal{M}_{n-1}^2+{K_{T,n}^2\theta_n^2}}\right.\\
		&\left. +2\mathcal{M}_{n-1} K_{T,n}\theta_n\cos\left( \phi_{n-1}-\phi_{n}+\zeta \right) \right)^{\frac{1}{2}},
	\end{aligned}
\end{equation}
\begin{equation}
	\normalsize
	\begin{aligned}
	\widehat{\phi}_{n-1}\! = \pi-\mathrm{atan2}
&\left\lbrace \mathcal{M}_{n-1}\cos\phi_{n-1}+K_{T,n}\theta_n\cos\left(\phi_n-\zeta\right)\right.\!,\\
&\left.\mathcal{M}_{n-1}\sin\phi_{n-1}+K_{T,n}\theta_n\sin\left(\phi_n-\zeta\right)\right\rbrace.
	\end{aligned}
\end{equation}
Hence, the coupled curvature of the ($n-1$)-th segment can be obtained as
\begin{equation}
	\normalsize
	\widehat{\kappa}_{n-1}=\frac{\widehat{\theta}_{n-1}}{s_{n-1}}.
\end{equation}
{Note that these coupled configurations are ``compression-responsive'', implying infinite solutions in the joint space unless with the bending stiffness $K_{T,k}$ being specified.} 



\section{Robot Description}
\label{robot description}
\begin{figure}[t!]
	\centering
	\includegraphics[width=3.3in]{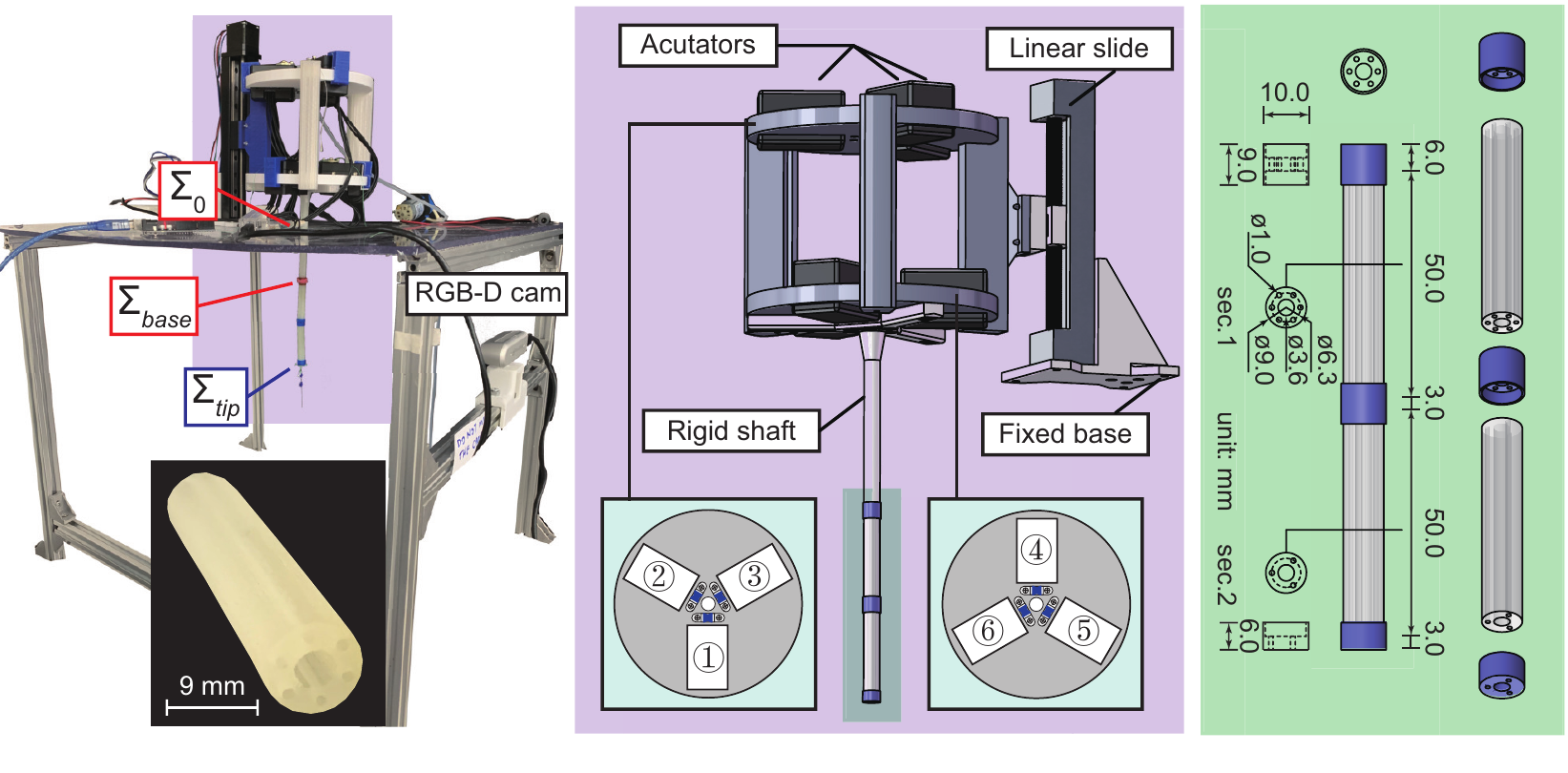}
	\caption{Design and Setup of the prototype. \textcircled{1}-$q_{1,1}$ \textcircled{2}-$q_{1,2}$ \textcircled{3}-$q_{1,3}$ \textcircled{4}-$q_{2,1}$ \textcircled{5}-$q_{2,2}$ \textcircled{6}-$q_{2,3}$. Young's modulus $E$ = 0.8 MPa. Original length $L_k$ = 50 mm.
	}
	\label{prototype}
\end{figure}

In this work, we evaluated the methodology using a two-segment CDSR prototype in \cite{unpublish}. The dimension and physical parameters of the robot are given in Fig. \ref{prototype}.
As shown in Fig. \ref{prototype}, for the second segment, three servo motors with actuation cables are installed with an offset of 180$\degree$, viz., $\zeta=\pi$. The manipulator is vertically mounted on a linear slide to form an insertion-retraction motion which is noted as $q_0$ and $q_0\geq 0$. Therefore, the actuator space can be expressed as
\begin{equation}
	\normalsize
	\bm{q}=\begin{bmatrix}
		{q_0}&q_{1,1}&q_{1,2}&q_{1,3}&q_{2,1}&q_{2,2}&q_{2,3}
	\end{bmatrix}^{\top},
\end{equation}
where $\bm{q}_{c}:=\begin{bmatrix}q_{1,1}&q_{1,2}&q_{1,3}&q_{2,1}&q_{2,2}&q_{2,3}\end{bmatrix}^{{\top}}$ is defined as the cable actuation. The cables we used were braided fish wire, which was relatively stiff in the axial direction (maximum 30 kg-force of tension) and thus they are assumed to be inextensible under the actuation of the selected servo motors (Dynamixel XM430-W350, ROBOTIS) with a capability of maximum 10 Nm of torque. Therefore, the tensile force of cables can be analogously directed as the motor actuation. The homogeneous transformation matrix of the tip with respect to a spatially-fixed world frame $\Sigma_0$ can be expressed as
\begin{equation}
		\mathbf{T}_{tip}^{0}\left(\bm{q} \right) 
		=\begin{bmatrix}
			\mathbf{I}_{3\times 3}&\mathbf{\mathcal{H}}\\\mathbf{0}_{1\times3}&1
		\end{bmatrix}\begin{bmatrix}
			\widehat{\mathbf{R}}_{1}^{0}&\widehat{\mathbf{x}}_{1}^{0}\\\mathbf{0}_{1\times3}&1
		\end{bmatrix}\begin{bmatrix}
			\widehat{\mathbf{R}}_{2}^{1}&\widehat{\mathbf{x}}_{2}^{1}\\\mathbf{0}_{1\times3}&1
		\end{bmatrix},
\end{equation}
where $\mathcal{H}=[0,0,q_0]^{\top}$ denotes the slide motion.


\section{Optimization-Based Constrained Motion Planning for Trajectory Tracking}
\label{method}

Given a linear velocity for the robot tip $\dot{\mathbf{x}}\in\mathbb{R}^{3}$ where $\dot{\mathbf{x}}=\dot{\mathbf{x}}^0_k$, the corresponding actuation velocity that drives that robot tip to reach the goal shall be inversely computed by
\begin{equation}
	\normalsize
	\dot{\bm{q}}=\mathbf{J}^{\dagger}\dot{\mathbf{x}} +\left(\mathbf{I}- \mathbf{J}^{\dagger}\mathbf{J}\right)\dot{\bm{q}}_{\mathcal{N}},
	\label{nullspace}
\end{equation}
where $\mathbf{J}\left(\bm{q} \right)\in\mathbb{R}^{3\times7}$ is a Jacobian matrix inferring a redundant robot, and ${\left( \cdot\right) }^\dagger$ denotes the right pseudo inverse operator as $\mathbf{J}^{{\top}}\left(\mathbf{J}\mathbf{J}^{{\top}} \right)^{-1} $. The second term projects the components of $\dot{\bm{q}}_{\mathcal{N}}$ to the null space of $\mathbf{J}$, with $\mathbf{I}$ being a 7-by-7 identity matrix. When the Jacobian matrix is updated in each time instance, the change in the tip position in an infinitesimal period of time $\Delta\mathbf{x}$ can be approximated based on the change of the actuation $\Delta\bm{q}$, such that
	$\Delta\bm{q}=\mathbf{J}^{\dagger}\Delta\mathbf{x}$.

The above equation may provide infinite solutions due to the system redundancy. However, it can be solved using some numerical approaches. One can solve the inverse kinematics of using a Jacobian-based pseudo-inverse method with a full rank $\mathbf{J}\left(\bm{q} \right)$. The other numerical approach utilizes the damped least-squares (DLS) inverse \cite{siciliano2010robotics}, which can avoid singularities with proper selection of the damping parameter $\lambda$, and provides a numerically stable method of selecting $\Delta\bm{q}$. In order to solve the needed actuation $\bm{q}$ for a desired path $\mathcal{X}_d=[\mathbf{x}_{d}^{(1)},...,\mathbf{x}_{d}^{(N)}]$ expressed by a finite set of discrete $N$ nodes, we compute the inverse kinematics (IK) by solving a convex least-square optimization problem as
\begin{equation}
	\normalsize
	\label{opt}
	\begin{aligned}
		\mathop{\arg\min}_{\Delta\bm{q}}\,  {\|\mathbf{J}\Delta\bm{q} - \Delta\mathbf{x}\|_2^2 + \lambda^2\| \Delta\bm{q} \|^2_2},
	\end{aligned}
\end{equation}
where $\lambda\in\mathbb{R}_{>0}$ is a non-zero damping constant. This quadratic optimization computes $\Delta\bm{q}$ that minimizes the error between the generated tip position and the desired tip position (first term), considering the feasible minimum motion in the actuation space (second term). It can generate one of the solutions which fulfills the requirement of tip path following, but the manipulator's motion may not be optimal. By exploiting the null space as introduced in (\ref{nullspace}), more sub-goals can be included to satisfy different constraints from the environment and on the robot. In this paper, we consider two important constraints for the motion planning.

%

\subsection{Constraint 1: Fixed Orientation End-Effector}
The control of the orientation of the soft robot's tip is crucial in many applications but has rarely been reported. However, the problem can be well-solved by adding the incremental change of tip orientation as an optimization term. Let the instantaneous change of tip orientation at the $j$-th point of the desired path $\mathcal{X}_d$ be represented by the Euler angles as 
$	\mathbf{\Omega}\left(\Delta\bm{q}_c^{(j) }\right) =\Delta\begin{bmatrix}\alpha^{(j)}&\beta^{(j)}&\gamma^{(j)}\end{bmatrix}^{\top}$.
The Euler orientation of the tip is only related to the cable actuation $\bm{q}_c \subseteq \mathcal{Q}^{3n}$. Hence, the optimization (\ref{opt}) can be reformulated as
\begin{equation}
	\normalsize
	\label{opt1}
	\begin{aligned}
	\mathop{\arg\min}_{\Delta\bm{q}}\,& \|\mathbf{J}\Delta\bm{q} \!-\!\Delta \mathbf{x}\|_2^2 \! + \!\sum_{i=1}^{3}\| \mathbf{\Omega}_i\left(\Delta\bm{q}_{c }\right)\!-\!\Delta\mathbf{\Omega}_i \|_2^2 \! +\! \lambda^2\| \Delta\bm{q}_c \|_2^2\\
	\mathrm{s.t.} \quad & \bm{A}\cdot\Delta\bm{q}\leq \bm{b}\quad\mathrm{and}\quad
	\bm{q}_{\min}\leq\bm{q}\leq\bm{q}_{\max},
	\end{aligned}
\end{equation}
with $\bm{A}=\textrm{diag}(10^{-3}, 1, 1, 1, 1, 1, 1)$ and $\bm{b}=0.01\cdot\textrm{ones}(7,1)$, both of which define the linear inequality constraints. The parameters are empirically selected with the rule that the slide motion shall be minimal compared to the manipulator's, otherwise, the slide (joint) motion will be significant, compared to the cables', as they share the same unit of mm. $\mathbf{\Omega}_i$ means the $i$-th row of the column vector, and $\Delta\mathbf{\Omega}_i$ is the desired change in tip orientation. Based on the work region of the robot tip, the bounded conditions are set as $\bm{q}_{\min}=-[0, 2, 2, 2, 2, 2, 2]^{{\top}}$,
$\bm{q}_{\max}=[60, 0, 0, 0, 0, 0, 0]^{{\top}}$.
Compared to (\ref{opt}), the additional term in (\ref{opt1}) minimizes the error between the generated task-space orientation change and the desired task-space orientation change under the linear inequality and bounded constraints.
 

\subsection{Constraint 2: Manipulator-Obstacle Collision Avoidance}
As introduced in (\ref{nullspace}), the null space motion enables the CDSR to perform tip trajectory tracking in an obstructed environment without any manipulator--obstacle collision.

\begin{figure}[t!]
	\centering
	\includegraphics[width=2.6in]{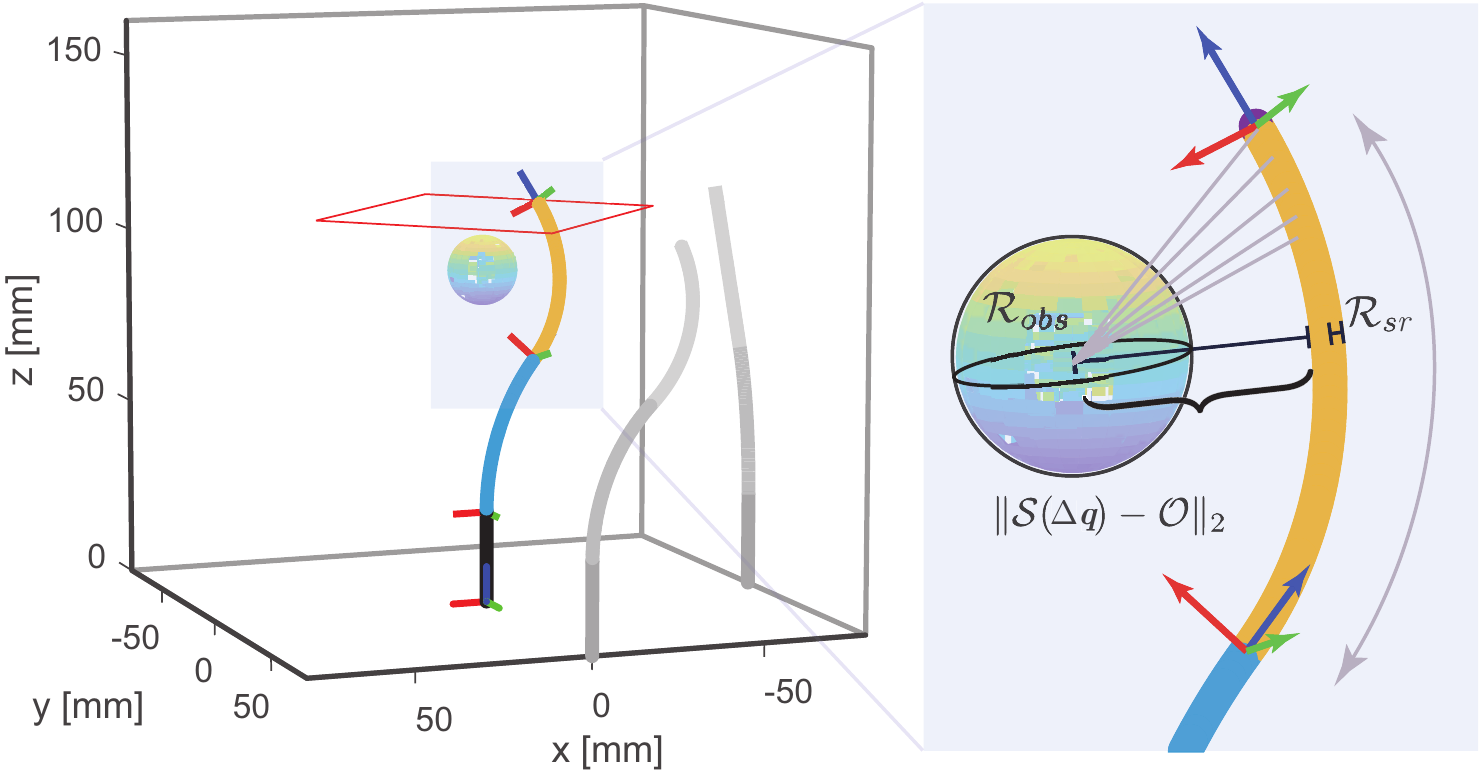}
	\caption{Interpretation of eq. (\ref{collision indicator}) for collision avoidance. One can iteratively compute the distance between the obstacle and each discrete point along the cylindrical robot surface.}
	\label{obs}
\end{figure}

For an obstacle with an arbitrary shape, a sphere, with a radius of $\mathcal{R}_{obs}$ and a spatial centroid of $\mathcal{O}$, can be defined to enclose the entire obstacle. The (cylindrical) radius of the CDSR is $\mathcal{R}_{sr}$, and therefore the critical distance between the virtual backbone of the CDSR and the obstacle centroid for collision is $\mathcal{R}_{obs}+\mathcal{R}_{sr}$. Thus, the critical distance at the moment of collision shall be defined as the minimal Euclidean distance that connects the obstacle centroid $\mathcal{O}$ to the critical point of the robot virtual backbone at a particular instant of actuation $\mathcal{S}\left(\Delta\bm{q} \right)$ as
\begin{equation}
\textrm{if}\;\;\min\|\mathcal{S}\left(\Delta\bm{q} \right) -\mathcal{O} \|_2
	\begin{cases}
	<\mathcal{R}_{obs}\!+\!\mathcal{R}_{sr}&\, \textrm{collision}\\
	=\mathcal{R}_{obs}\!+\!\mathcal{R}_{sr}&\, \textrm{critical dist.}\\
	>\mathcal{R}_{obs}\!+\!\mathcal{R}_{sr}&\, \textrm{no collision},\\
	\end{cases}
\end{equation}
where $\min\left(\cdot\right)$ returns the minimum of $\left( \cdot\right)$. Therefore, as shown in Fig. \ref{obs}, the collision avoidance can be regulated as the minimization of
\begin{equation}
	\normalsize
	G\left(\Delta\bm{q} \right) =\frac{\mathcal{R}_{obs}+\mathcal{R}_{sr}}{\min\|\mathcal{S}\left(\Delta\bm{q} \right) -\mathcal{O} \|_2 }.
	\label{collision indicator}
\end{equation}
Noted that $G\left(\Delta\bm{q}\right)\rightarrow0 $ refers to the cases without possible collision, i.e., $\min\|\mathcal{S}\left(\Delta\bm{q} \right) -\mathcal{O} \|_2\gg{\mathcal{R}_{obs}+\mathcal{R}_{sr}}$. In this regard, equation (\ref{opt}) can be revised as
\begin{equation}
	\normalsize
	\label{opt2}
	\begin{aligned}
		\mathop{\arg\min}_{\Delta\bm{q}}\, & {\|\mathbf{J}\Delta\bm{q} - \Delta\mathbf{x}\|_2^2 +\eta \|G\left(\Delta\bm{q} \right)\|_2^2 + \lambda^2\| \Delta\bm{q}_c \|_2^2}  \\
		\mathrm{s.t.} \quad & \bm{A}\cdot\Delta\bm{q}\leq \bm{b}\quad\mathrm{and}\quad
		\bm{q}_{\min}\leq\bm{q}\leq \bm{q}_{\max},
	\end{aligned}
\end{equation}
where $\eta$ is a positive constant to weight the sensitivity of collision avoidance. The linear inequality and bounded values still follow the setting in Constraint 1.

\section{Simulation}
\label{simulation}
This section will demonstrate the simulated results of the constrained motion of a two-segment CDSR for tracing the pre-defined {paths without considering the time prescription}, with the additional requirement being fulfilled. 
The simulations were written in Matlab using the parameters of our prototype. {The animation was built based on the plotting of a number of 3-D points along the robot body by computing the compressible curvature forward kinematics using the inverse solution that being solved from the optimization.} Codes are available online\footnote[1]{URL: \href{https://github.com/samlaipolyu/ncc\_motion\_planning}{{{https://github.com/samlaipolyu/ncc\_motion\_planning}}}}. Please refer to the \texttt{README.md} files for implementation. All simulations were performed using a PC with an Intel Core i7-8750H CPU @2.20GHz 16GB RAM.

\begin{figure}[t]
	\centering
	\includegraphics[scale=0.32]{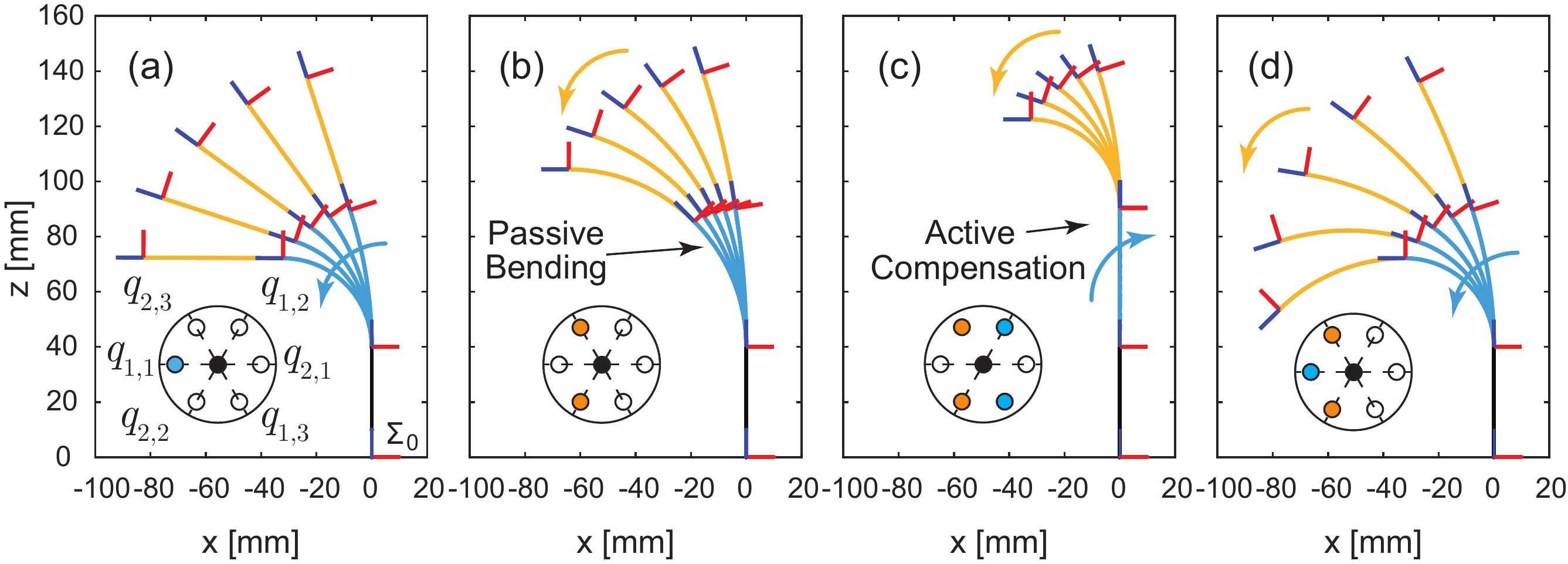}
	\caption{Coupling effect of a two-segment CDSR in simulation. (a) Only the proximal segment is bent; (b) only the distal segment is bent; (c) both segments were bent, but the proximal bending compensated the distal bending; (d) both segments were bent in the same direction.}
	\label{bending_demo}
\end{figure}

The basic bending motion of the CDSR using the improved  compressible curvature model is shown in Fig. \ref{bending_demo}. It can be seen that, due to the coupling effect, actuating the distal segment would result in a passive bending for the proximal segment.
\begin{algorithm}[t!]
	\caption{\small Optimization of constrained motion of CDSR}
	\label{algorithm}
	\small
	\KwIn{Desired path/traj. $\mathcal{X}_d=[\mathbf{x}_{d}^{(1)},...,\mathbf{x}_{d}^{(N)}]\in\mathbb{R}^{3\times N}$, optimization para. and constraints, desired orientation $\mathbf{\Omega}_d\in\mathbb{R}^{3\times N}$ or obstacle para. $\mathcal{R}_{obs}$, $\mathcal{O}$}
	\KwResult{Actuation $\bm{q}\in\mathbb{R}^{7\times N}$ for the contrained motion}
	\For{$i=1:N $}
	{
		\While{$\|\Delta\mathbf{x}\|_2>{threshold}$}
		{
			Query $\bm{q}_{current}\in\mathbb{R}^{7\times1}$\;
			Calculate $\mathbf{x}_{tip}^0=forwardKine\left( \bm{q}_{current}\right)$\;
			Calculate $\Delta\mathbf{x}=\mathbf{x}_{d}^{(i)}-\mathbf{x}_{tip}^0$\;
			Compute $\Delta\bm{q}\leftarrow\arg\min_{\Delta\bm{q}} \{ objFun\left(\Delta\mathbf{x} \right) \}$ from (\ref{opt1}) or (\ref{opt2})\;
			Set $\bm{q}_{current}\leftarrow\bm{q}_{previous}+\Delta\bm{q}$\;
			Set $\bm{q}_{previous}\leftarrow\bm{q}_{current}$\;
		}
		Set $\bm{q}^{(i)}\leftarrow\bm{q}_{current}$;\
	}
\end{algorithm}
\subsection{Simulation 1: Fixed Orientaiton End-Effector}
\label{simulation.para.1}
The path tracking tasks using a fixed orientation tip were simulated based on the compressible curvature model. The related demos are available in the supplementary video. The simulation implemented equation (\ref{opt1}) with the initial value of $\Delta\bm{q}_{init}=-10^{-2}[-0.01, 1, 2, 1, 2, 1, 2]^{\top}$ and an empirically-selected damping coefficient\footnote[2]{Larger $\lambda$ will result in a ``stiffer'' motion. Readers may try different values using our code according to the practical need.} of $\lambda^2=0.1$. The trial path were represented by a finite number of nodes along its path using parabolic functions. 
The tracking time can be controlled and predicted based on $N$ and the computational time of each interval. The control algorithm is shown in Algorithm \ref{algorithm}. The goal is to use the optimized inverse solution to implement the CDSR such that it can follow a pre-defined path with a constrained motion. {The time required to compute the solution using the PC with the given specification is about 0.4 second per for-loop using pure Matlab script. It can be further boosted to an update rate of 60 Hz when being converted to C.}

\begin{figure*}[t!]
	\centering
	\includegraphics[width=7.1in]{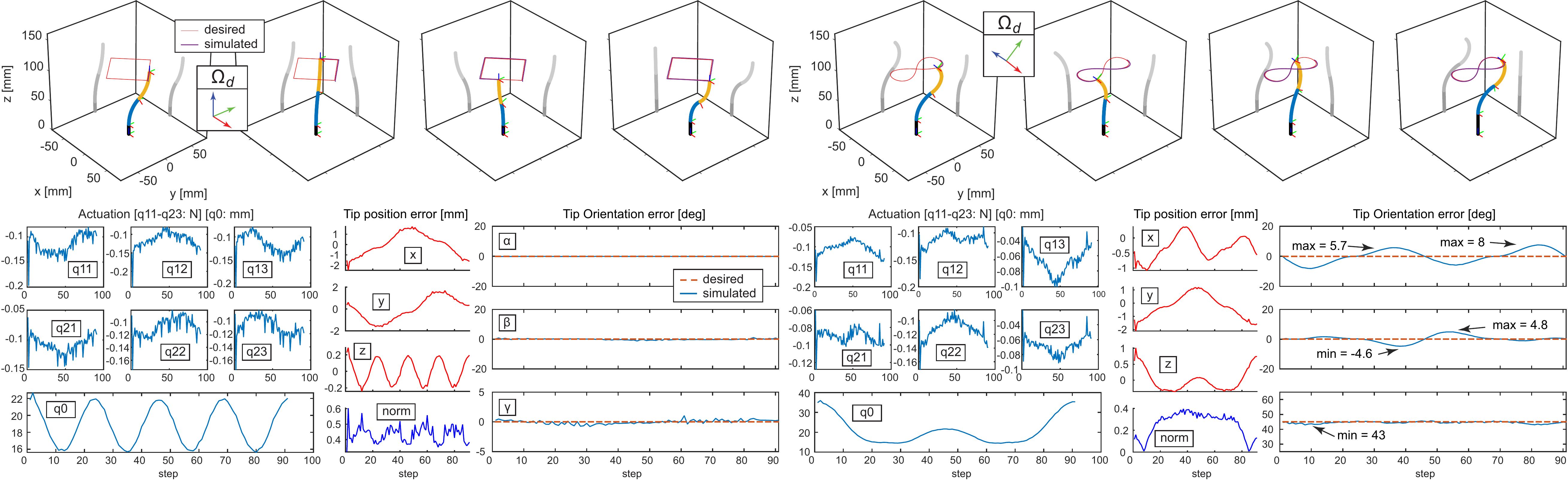}
	\caption{Simulation 1: tracking paths with the orientation of the end-effector fixed. Left: tracking a square with the tip pointing ``upward" with respect to the CDSR. Right: tracking an eight with its tip tilted by 45$\degree$. The corresponding cable actuation, tip position error, and tip orientation error are given$^{\textrm{a}}$.}
	\label{simulation1}
\end{figure*}

The simulation results of two examples are shown in Fig. \ref{simulation1}. The left part of the figure demonstrates the tracking performance of a planar square path (with $N=91$ nodes) with its fixed orientation end-effector. The desired tip orientation was set as $\mathbf{\Omega}_d=[0, 0, 0]^{\top}$, i.e., intuitively, pointing ``upward" with respect to the base frame, forming a vertical tip pose throughout the path tracking process. With the desired path and orientation predefined, the proposed DLS IK solver computes the needed cable actuation in the form of tensile force ($\bm{f}<0$) and the motion of linear slide $q_0$ to compensate the change of $z_{tip}^0$ for the desired working surfaces. Then, we input the actuation to the simulator to visualize the move using forward kinematics. The results indicate that CDSR motion generated from the computed actuation can satisfy the tip positioning within an absolute error of 0.8 mm. The orientation can be well aligned as desired within a 1-degree error.

The tip orientation with respect to the world frame can be assigned and maintained as much as possible. The right part of Fig. \ref{simulation1} shows the simulation result of tracking an ``8"-shaped path ($N=91$ nodes), with a desired orientation defined as $\mathbf{\Omega}_d=[0, 0, 45]^{\top}$ degrees. The computed tip position error norm is within 0.6 mm. The error in the tip pointing angle $\gamma$ is within 2$\degree$. The other two Euler angles, $\alpha$ and $\beta$, are fluctuating but within minor angular errors. Another two examples with more complex paths are given {in the supplementary}. The accuracy of maintaining the orientation depends on the magnitude of the tilted angle: higher accuracy could be expected when the tip is completely ``vertical''. The method is also applicable in a changing orientation $\mathbf{\Omega}_d$.

\subsection{Simulation 2: Manipulator-Obstacle Collision Avoidance}
The related demos are available in the supplementary video. The simulation implemented equation (\ref{opt2}) with the same $\Delta\bm{q}_{init}$ and $\lambda^2$ in Sec. \ref{simulation.para.1}. Regarding the obstacle setting, we consider a sphere located at $\mathcal{O} = [-10, -30, 90]^{\top}$ mm with respect to the world frame $\Sigma_0$, with a radius of $\mathcal{R}_{obs}=7.5$ mm. The collision avoidance sensitivity is $\eta=0.1$. The CDSR shall complete the path tracking task accurately but avoid any manipulator-obstacle collisions.
\begin{figure*}[t!]
	\centering
	\includegraphics[width=6.85in]{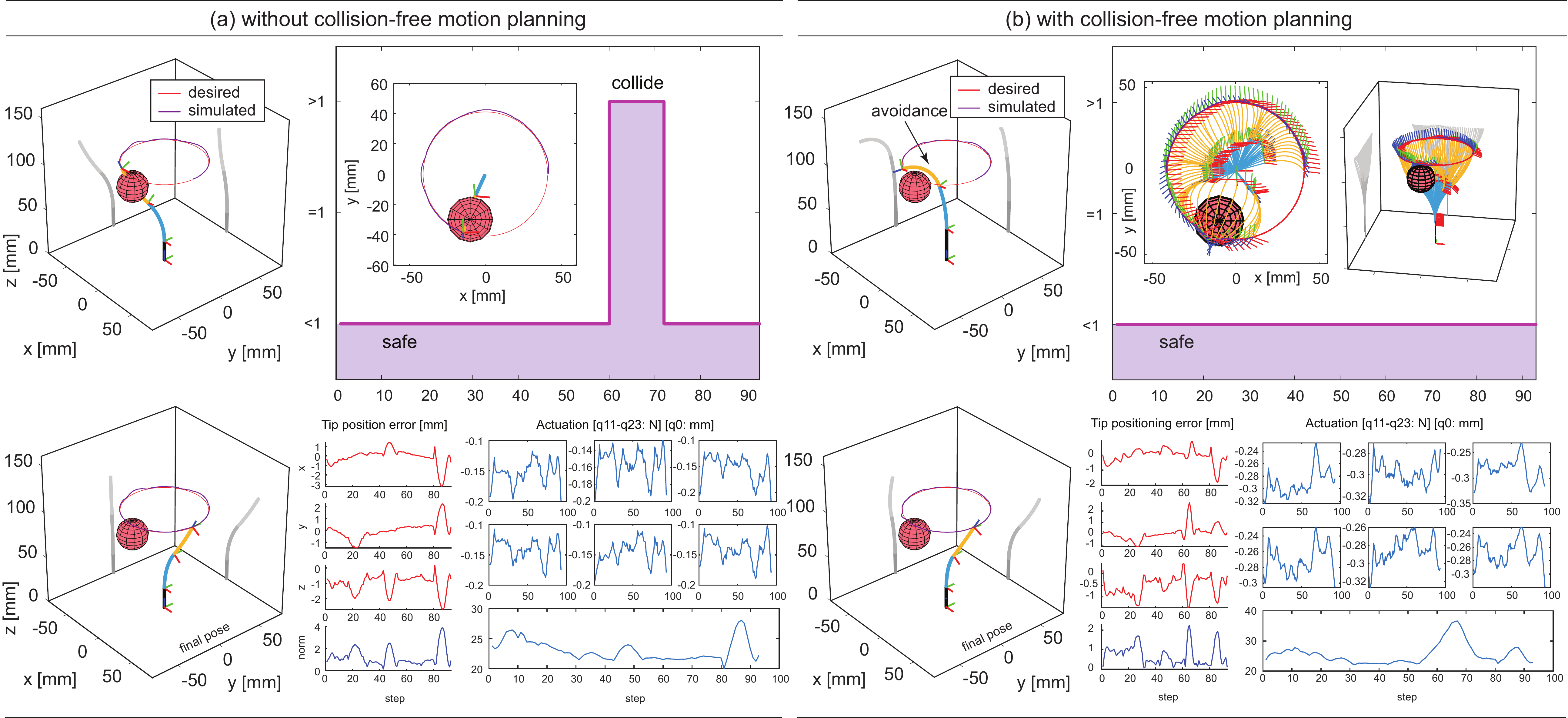}
	\caption{Simulation 2: tracking paths in a constrained environment, where (a) the CDSR would collide into the obstacle during the tracking process. (b) With proper motion planning, the CDSR can avoid such collision while keeping a satisfactory path tracking accuracy. All conditions are exactly the same except the with or without collision-free motion planning. The collision indicator is given by (\ref{collision indicator}). {More results are available in the supplementary.}}
	\label{simulation2}
\end{figure*}
To demonstrate the effectiveness of our method, as shown in Fig. \ref{simulation2}-a, {we provide a circular path with prescribed time, $\mathbf{x}_{d}^{(t)}=[41\cos\left(t \right), 41\sin\left(t \right), 110]^{\top}$ mm with $t=0:2\pi$ sec in an $\frac{\pi}{45}$ interval}, which would have had a collision with the robot body under normal IK solution from (\ref{opt}). By applying (\ref{opt2}) to algorithm \ref{algorithm}, the solver can provide a null space solution such that the CDSR can adjust its body motion and dodge the obstacle, while the tip keeps tracing the planned path (Fig. \ref{simulation2}-b).

\section{Experiment}
\label{expriment}

\subsection{Model Verification}
\label{Model Verification}
\begin{figure}[t!]
	\centering
	\includegraphics[width=3.3in]{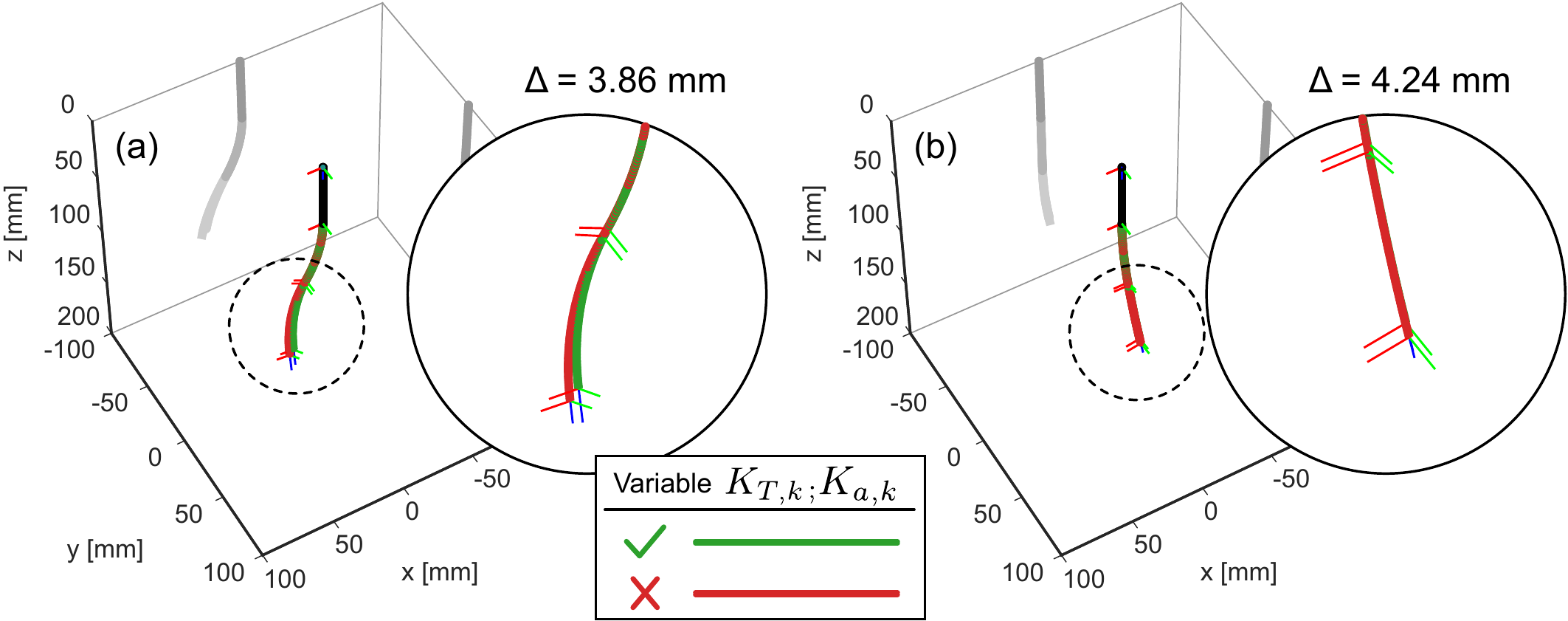}
	\caption{Comparison on compressible-associated $K_{a,k}$ and $K_{T,k}$ versus the constant $K_{a}$ and $K_{T}$ based on simulation. The variation will be more important on the use of softer materials and longer robots.}
	\label{mv1}
\end{figure}
{Based on the numerical simulation, we verified the differences on tip positioning between using the compressible-associated $K_{a,k}$ and $K_{T,k}$ versus the constant $K_{a}$ and $K_{T}$. Here we provide two examples in Fig. \ref{mv1}. We first employed a random set of actuation to the simulator and obtained a proposed robot pose (Green) using $K_{T,k}\left(s_k \right)$. Then, with the same actuation, and setting the bending stiffness to be a constantly defined $K_{a}$ and $K_{T}$, we could obtain a deviated robot pose (Red). Based on our 100 mm-long prototype, such tip positioning norm difference can be up to about 5 mm. {A larger difference will be expected when using softer materials or applying on longer robots}. The results indicate the importance of considering the axial compression when we were modeling a soft material-based actuator.}

\begin{figure}[t!]
	\centering
	\includegraphics[width=3.2in]{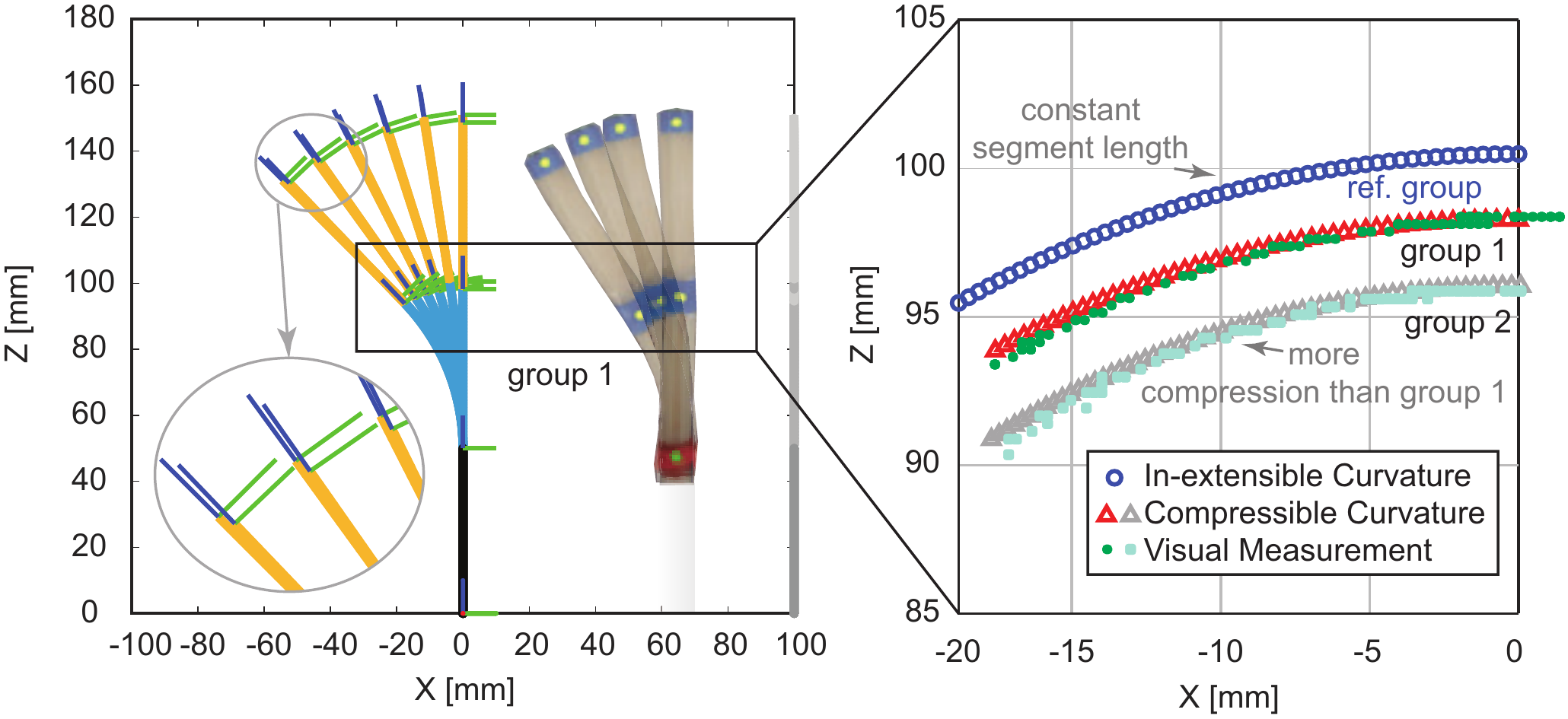}
	\caption{Verification of the compressible curvature model on a soft robot. Originated at the same point, pretension were added in group 1 and group 2 to shorten the segment length.}
	\label{validation}
\end{figure}

The compressible curvature model on a soft segment was experimentally verified. As depicted in Fig. \ref{validation}, a planar bending test was conducted on the proximal segment of a CDSR. Due to the material softness, the segments would undergo an axial compression when the cables were actuated. Here we use an inextensible PCC model that excludes the arc length variation when the robot is in motion, i.e., $s_k\left(\bm{f}\neq0\right)=L_k$ as the baseline. Note that the PCC model should be fitted to when $s_k(\bm{f}=0)=L_k$. {In Fig. \ref{validation}, group 1 and group 2 denote the total arc length of 98 and 96 mm, respectively.} The result indicates that the inextensible PCC model for general continuum robots may not be satisfactory enough for soft-bodied robots. However, our compressible curvature model can better predict the tip position with given actuation inputs.

\subsection{Constrained Motion Planning}
Experiments were conducted to evaluate the performance of the proposed motion planning approach for a two-segment CDSR prototype based on the compressible curvature model. Since the tensile force that applying on the inextensible cables can be analogously viewed as the cable displacement, we experimentally scaled the proportion between the applied tensile force and cable displacement, i.e., $\bm{q}_c=\mu\bm{f}$, where $\mu=18$ mm/N based on the tuned cable tension. 
To produce a smooth input for the motor, a moving median filter was used to reduce the periodic trends and outliers in the solution. 
A bench-top RGB-D camera (RealSense D435, Intel) was used to retrieve the features of a labeled CDSR as shown in Fig. \ref{prototype}, and transform the data to the world frame coordinate.

\begin{figure}[t!]
	\centering
	\includegraphics[scale=0.27]{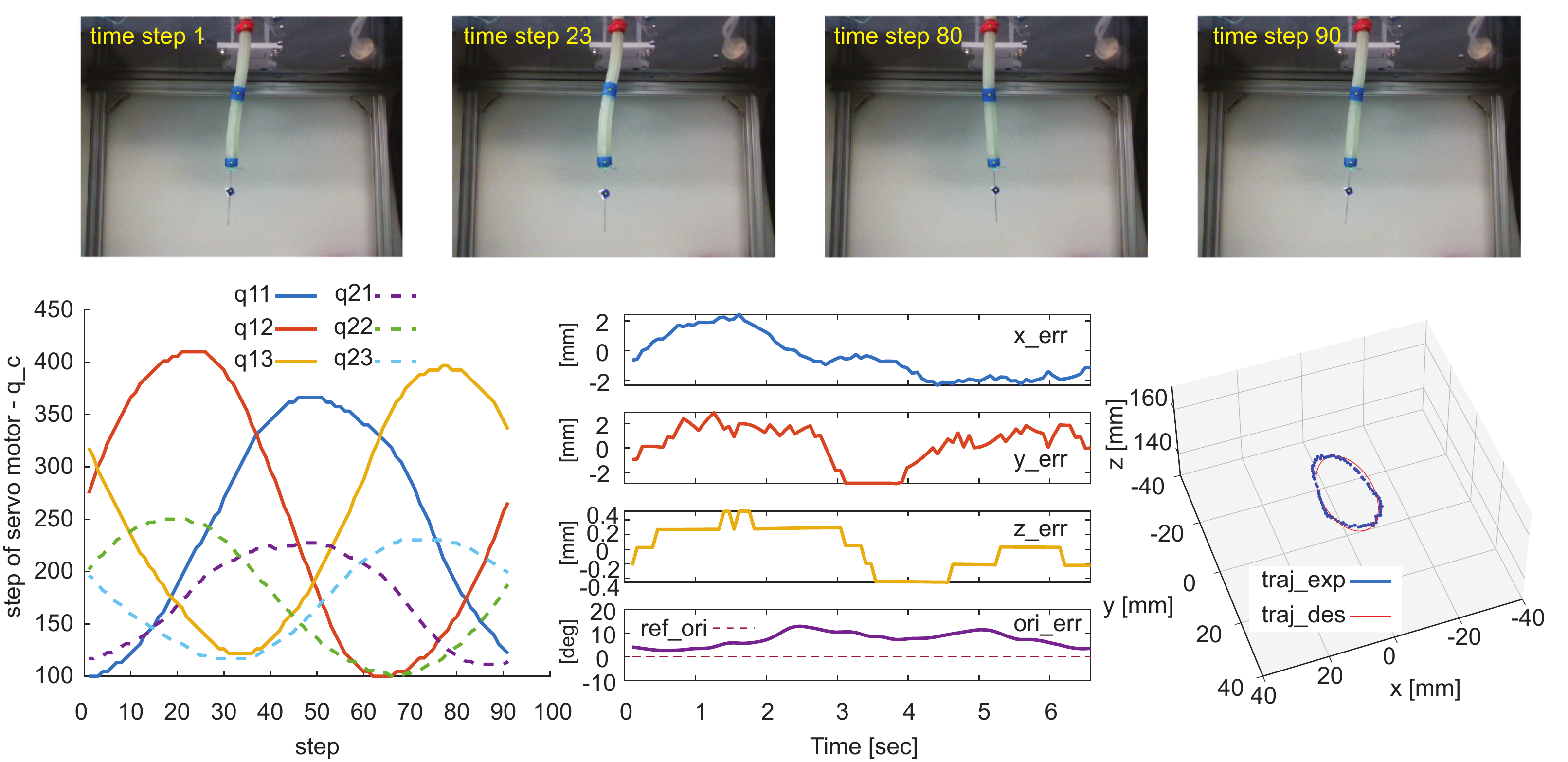}
	\caption{Experiment of tracking an oval with the tip being vertical to the ground (tilt angle = 0$\degree$). Snapshot by time sequence; actuation input $\bm{q}_c$; tip positioning error on 3 axes and orientation error; and overall tracking performance.}
	\label{exp_1}
\end{figure}

\begin{figure}[t!]
	\centering
	\includegraphics[scale=0.27]{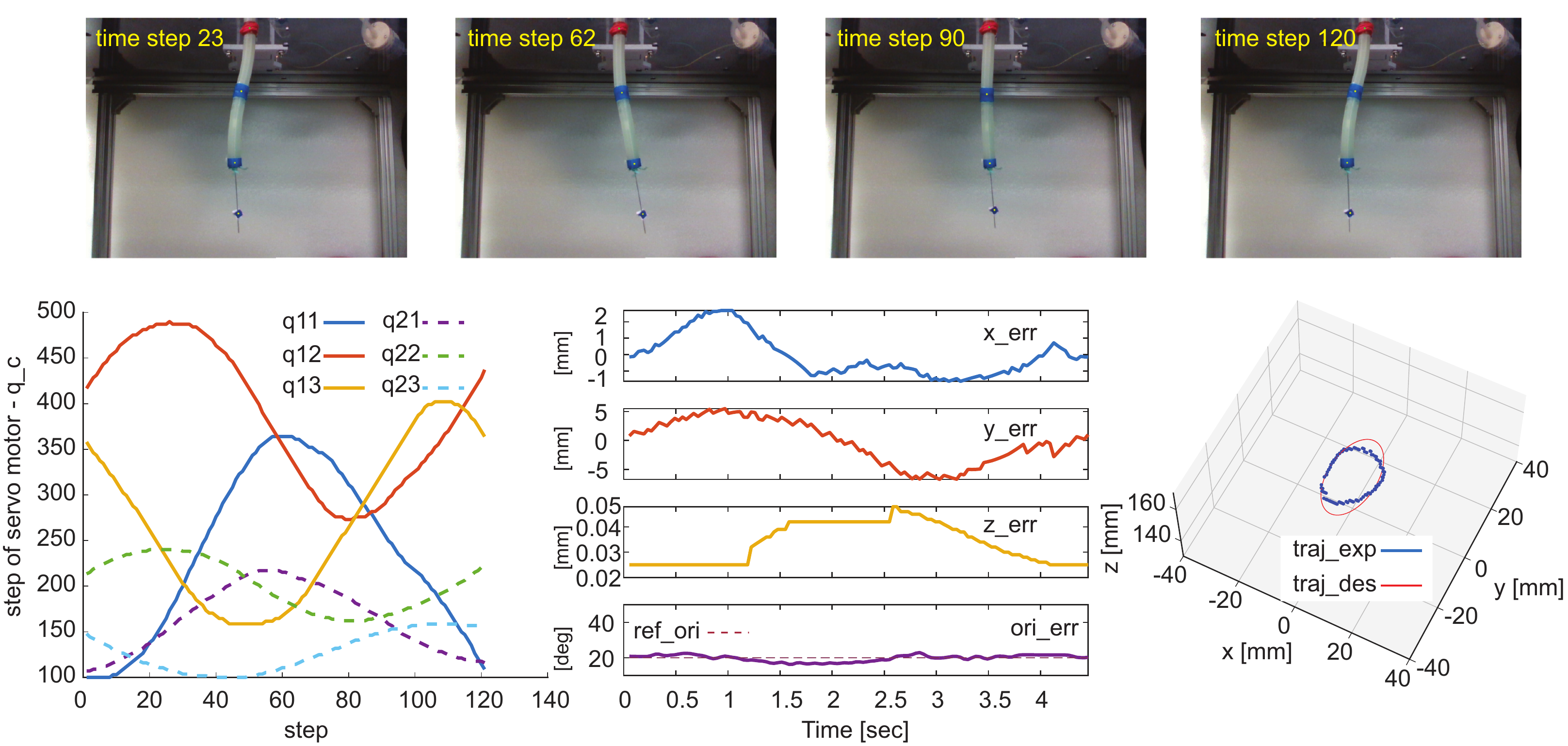}
	\caption{Experiment of tracking an oval with the tip being tilted to the ground (tilt angle = 20$\degree$).}
	\label{exp_2}
\end{figure}

The performance of trajectory tracking under the constraint 1 (fixed orientation) have been tested. {We used parametric functions of time step to represent the trajectories.} We verified that the computed input can produce constrained manipulator motions as desired with satisfactory accuracy. Fig. \ref{exp_1} shows the results in the case where the end-effector was required to trace an oval with its tip orientation being vertical to the ground. The visual measurement indicates an axial error within 2 mm and an orientation error within 10$\degree$. Fig. \ref{exp_2} shows the situation when a 20$\degree$ tilt angle was assigned to the CDSR under the same path. Even though the positioning accuracy at the long end of the elliptical path was not as good as in the vertical case as the manipulator attempted to flex itself to maintain the certain tip orientation along with the tracking, the result was satisfactory in fulfilling the fixed orientation requirement.

\begin{figure}[t]
	\centering
	\includegraphics[scale=0.27]{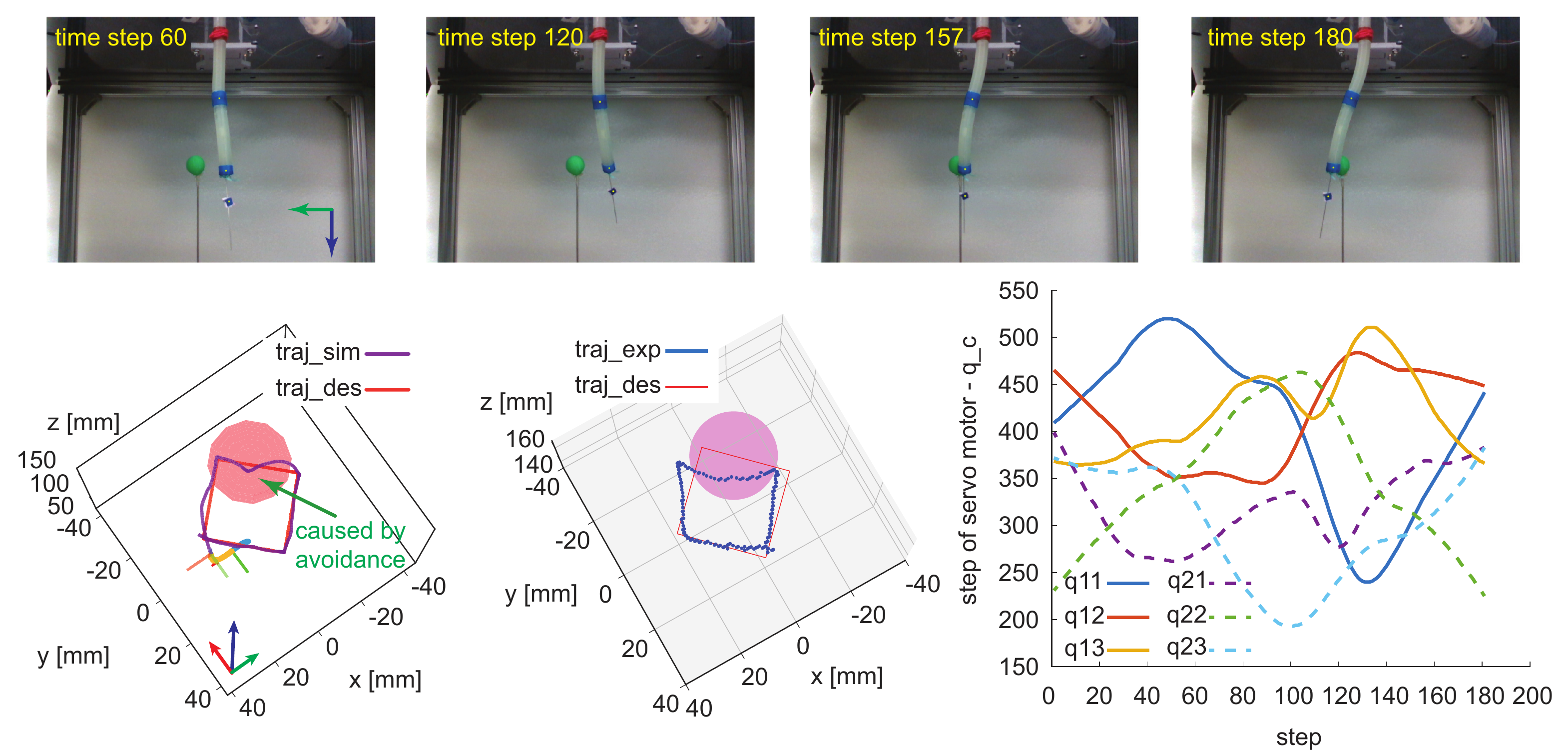}
	\caption{Experiment of tracking a square in an obstructed environment with collision-free motion planning of the soft manipulator.}
	\label{exp_3}
\end{figure}

The performance of trajectory tracking under constraint 2 (collision avoidance) has also been tested on the prototype. A $\diameter$15 mm spherical obstacle was located at $[-10, -15, 95]$ mm in $\Sigma_0$, which would have had blocked the robot motion in tracking the square trajectory. By deploying the proposed motion planning method, the computed inverse solution was able to drive the CDSR to avoid collision with the desired tip path being tracked (Fig. \ref{exp_3}). The experiment shows that the tracking error was within 5 mm, which was overall satisfactory. However, it should be noted that, aligning with the simulation result, the error of tracking would be enlarged near the critical point of collision due to the deliberate motion for the avoidance. This can be treated as a trade-off for the collision-free motion depending on the selection of $\eta$.

\section{Conclusions}
\label{summary}
In this work, we demonstrated a constrained motion planning approach incorporated with the compressible curvature modeling for a multisegment cable-driven soft robot under the segment coupling effect. {The robot modeling includes the undesirable segment shortening due to the material elasticity and cable-driven mechanism. Based on the mechanics of the cable and soft body, the decoupled mapping among spaces of the actuator, configuration, and task were derived.} On top of that, we developed an optimization-based motion planning algorithm to extend the controllability of a redundant soft robot for tip trajectory tracking in constrained conditions, including with a fixed orientation tip, and with the avoidable manipulator--obstacle collision. Based on a two-segment robot prototype, numerical simulation and experiments were carried out to confirm and evaluate the {explicitly-formulated model} and the proposed algorithm for tip trajectory tracking tasks. Our method may be generalized for similar multisegment cable-driven soft robots in constrained motion planning.

However, force-related interaction has been excluded in the current method. Future effort may be made in studying the variable-stiffness motion planning by exploiting the soft body compression which varies the soft body's stiffness.

\bibliographystyle{IEEEtran}
\bibliography{preprint}





\end{document}